\RequirePackage{amsmath}%
\RequirePackage{amssymb}%
\documentclass[runningheads]{llncs}
\usepackage[T1]{fontenc}
\usepackage{microtype}
\usepackage{graphicx}
\usepackage{hyperref}
\usepackage{color}

\usepackage[labelfont=bf,labelsep=period]{caption}
\usepackage[labelformat=parens,subrefformat=parens]{subcaption}
\usepackage[capitalise,noabbrev]{cleveref}

\usepackage{numprint}
\usepackage{paralist}

\usepackage{booktabs}%
\usepackage{tabularx}
\usepackage{dcolumn}

\usepackage{comment}
\begin{document}
\title{Comparing Deep Reinforcement Learning Algorithms in Two-Echelon Supply Chains}
\titlerunning{Comparing DRL Algorithms in Two-Echelon Supply Chains}
\author{Francesco Stranieri\thanks{Corresponding author. E-mail: \email{francesco.stranieri@polito.it}.}\inst{1,2}\orcidID{0000-0002-5366-8499} \and
\\Fabio Stella\inst{1}\orcidID{0000-0002-1394-0507}}
\authorrunning{F. Stranieri and F. Stella}
\institute{University of Milan-Bicocca, Milan MI 20125, Italy \and
Polytechnic of Turin, Turin TO 10129, Italy}
\maketitle              % typeset the header of the contribution
\begin{abstract}
In this study, we analyze and compare the performance of state-of-the-art deep reinforcement learning algorithms for solving the supply chain inventory management problem. 
This complex sequential decision-making problem consists of determining the optimal quantity of products to be produced and shipped across different warehouses over a given time horizon.
In particular, we present a mathematical formulation of a two-echelon supply chain environment with stochastic and seasonal demand, which allows managing an arbitrary number of warehouses and product types.
Through a rich set of numerical experiments, we compare the performance of different deep reinforcement learning algorithms under various supply chain structures, topologies, demands, capacities, and costs.
The results of the experimental plan indicate that deep reinforcement learning algorithms outperform traditional inventory management strategies, such as the static (s, Q)-policy.
Furthermore, this study provides detailed insight into the design and development of an open-source software library that provides a customizable environment for solving the supply chain inventory management problem using a wide range of data-driven approaches.

\end{abstract}
\section{Introduction}\label{section1}
Supply chain inventory management (SCIM) is a \textit{sequential decision-making problem} consisting of determining the optimal quantity of products to produce at the factory and to ship to different distribution warehouses over a given time horizon.
As evidenced by the helpful roadmap of \cite{Boute2022}, deep reinforcement learning (DRL) algorithms are rarely applied to the SCIM field, although they can be used to develop near-optimal policies that are difficult, or impossible at worst, to achieve using traditional methods.
Indeed, the uncertain and stochastic nature of products demand, as well as lead times, represent significant obstacles for mathematical programming approaches to be effective, with specific reference to those cases where the modeling of SCIM's entities is reasonable, for example, assuming a finite capacity of warehouses \cite{deKok2018}.

Regarding the DRL algorithms that have been currently applied to tackle the SCIM problem, we found that they suffer the following \textit{limitations}:
\begin{inparaenum}[i)]
    \item given a supply chain \textit{structure} (e.g., divergent \footnote{In a \textit{linear} supply chain, each participant has one predecessor and one successor;
    in a \textit{divergent} supply chain, each has one predecessor but can have multiple successors, while the opposite is true in a \textit{convergent} supply chain.
    Finally, in a \textit{general} supply chain, each participant can have several predecessors and several successors.}
    two-echelon \footnote{A supply chain can include multiple stages, called formally \textit{echelons}, through which the stocks are moved to reach the customer. 
    When the number of echelons is greater than one, we refer to a \textit{multi-echelon} supply chain.}), no DRL algorithm has been deeply tested with respect to different \textit{topologies} (i.e., by changing the number of warehouses);
    \item no extensive experiments have been performed on the same supply chain structure by varying different \textit{configurations} (e.g., demands, capacities, and costs);
    \item no extension has been proposed for \textit{comparing different DRL algorithms} and determining which one is more appropriate for a particular supply chain topology and configuration, as suggested by \cite{Alves2020,Boute2022}.
\end{inparaenum}

Furthermore, relevant aspects of the SCIM problem have not yet been addressed efficiently \cite{Yan2022}, for example:
\begin{inparaenum}[i)]
    \item the \textit{sequence of events} required to reproduce and validate a simulation model is not always well-defined or given.
    Hence, making available a consistent and universal open-source SCIM environment can improve reusability and reproducibility, especially if implemented with standard APIs (like those of OpenAI Gym \footnotemark).
    In this way, it is also possible to import DRL algorithms from reliable libraries and focus solely on their fine-tuning, instead of developing them from scratch;
    \item DRL algorithms are typically compared with some standard \textit{static reorder policies}.
    However, their performances are not always compared with those achieved by an oracle, i.e., a baseline who knows the optimal action to take a priori, thus making it difficult to evaluate the DRL effectiveness in real-world environments (the only paper in which an oracle is introduced is \cite{hubbs2020or});
    \item none of the DRL papers available in the specialized literature considers a \textit{multi-product approach}, whereas it has been considered relating to other solution methods \cite{Yan2022}.
    Considering more than one product type increases the dimensionality and complexity of the problem, consequently requiring an efficient implementation of the SCIM environment and DRL algorithms.
\end{inparaenum}

This paper makes the following \textit{contributions} to the SCIM decision-making problem:
\begin{itemize}
  \item Design and formulation of a stochastic and divergent two-echelon SCIM environment under seasonal demand, which allows an arbitrary number of warehouses and product types to be managed.
  \item Comparison of a set of state-of-the-art DRL algorithms in terms of their ability to find an optimal policy, i.e., a policy which maximizes the SCIM's profit as achieved by an oracle.
  \item Evaluation of performances achieved by state-of-the-art DRL algorithms and comparison to a static reorder policy, i.e., an ($s$, $Q$)-policy, whose optimal parameters have been set through a data-driven approach.
  \item Design and run of a rich experimental plan involving different SCIM topologies and configurations as well as values of hyperparameters associated with DRL algorithms'. \footnotetext{The OpenAI Gym library is available on \url{https://www.gymlibrary.dev}.}
  \item Design and development of an open-source library for solving the SCIM problem \footnote{Our open-source library is available on \url{https://github.com/frenkowski/SCIMAI-Gym}.}, thus embracing the open science principles and guaranteeing reproducible results.
\end{itemize}

The rest of the paper is organized as follows:
\cref{sec:section2} is devoted to introducing and providing main reinforcement learning (RL) definitions and notation, also highlighting how RL approaches have dealt with the SCIM problem;
in this section, we also describe the state-of-the-art DRL algorithms and how they have been used to address the SCIM problem.
\cref{sec:section4} describes the main methodological contributions of this paper.
The rich experimental plan is then reported in \cref{sec:section5}, while the results of numerical experiments are presented in \cref{sec:section6}.
Lastly, discussions and conclusions are given in \cref{sec:section7}.

\section{Literature Review} \label{sec:section2}
While reinforcement learning has recently achieved remarkable results in the field of artificial intelligence, mainly when applied to video games and gaming in a more general sense \cite{Mnih2015,Silver2017,Vinyals2019}, its deployment in industrial settings has been less extensive.
Despite RL proving to be effective in solving complex sequential decision-making problems, its translation into industrial use cases is still emerging, devising a concrete opportunity for further explore its potentialities \cite{Yan2022}.

\begin{figure}[ht]
    \centering
    \includegraphics[width=0.65\textwidth]{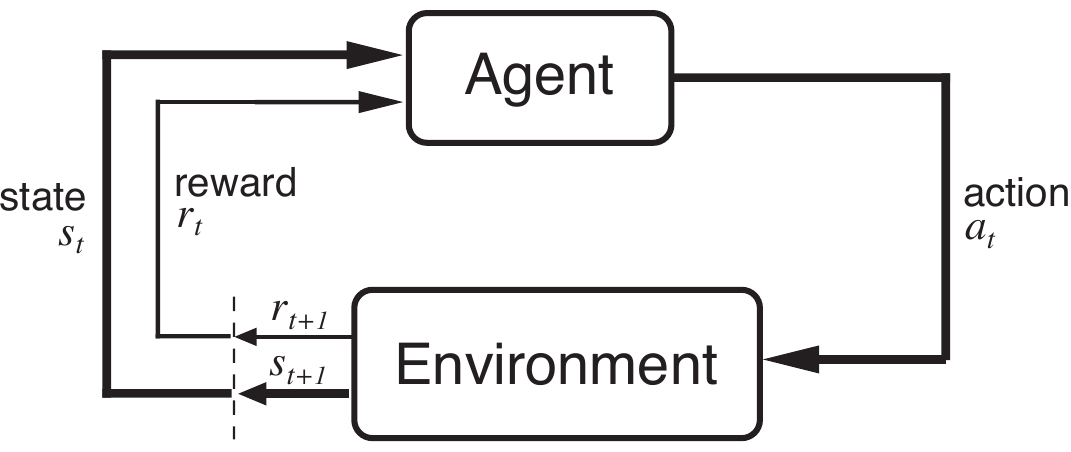}
    \caption{Agent-environment interface in an MDP (taken from \cite{sutton2018reinforcement}).}
    \label{fig:rl}
\end{figure}

Essentially, RL adopts the Markov Decision Process (MDP) framework to represent the interactions between a learning agent and an environment \cite{sutton2018reinforcement}.
As shown in \cref{fig:rl}, at each time step $t$, the agent observes the current state of the environment, $S_{t} \in \mathcal{S}$, chooses an action, $A_{t} \in \mathcal{A}(S_{t})$, and obtains a reward, $R_{t+1} \in \mathcal{R} \subset \mathbb{R}$; 
then, the environment transitions into a new state, $S_{t+1}$.
The goal of RL is thus to find an optimal policy, $\pi_{*}: \mathcal{S} \rightarrow \mathcal{A}$, that maximizes the \textit{expected discounted return}, $G_{t} = \sum_{k=t+1}^{T} \gamma^{k-t-1} R_{k}$, where $0 \leq \gamma \leq 1$ is a hyperparameter called \textit{discount rate}.

One of the most common approaches for solving the SCIM problem through RL algorithms turns out to be Q-learning.
This approach is based on a tabular and temporal-difference (TD) algorithm that learns how to determine the \textit{value} of an action $A_{t}$ in a state $S_{t}$, referred to as the Q-value, in accordance with the following update rule: $Q\left(S_{t}, A_{t}\right) \leftarrow Q\left(S_{t}, A_{t}\right)+\alpha \delta_{t}$, where $0 \leq \alpha \leq 1$ is a hyperparameter called \textit{learning rate}, and $\delta_{t} = \left[R_{t+1}+\gamma \max _{a} Q\left(S_{t+1}, a\right)-Q\left(S_{t}, A_{t}\right)\right]$ is the TD error.
Q-values of each state-action pair are stored in a table, known as Q-table, where each state is represented by a row and each action by a column.
Through the Q-learning algorithm, Q-values associated with each state-action pair are estimated and, once convergence has been achieved (which is guaranteed under certain conditions \cite{Jaakkola1994}), an optimal policy can be easily obtained by identifying, for each state, the action with the highest Q-value, that is, $\pi_{*}(S_{t})=\arg \max _{a} Q\left(S_{t}, a\right)$.

In \cite{Chaharsooghi2008}, which is one of the most cited RL articles about SCIM, the authors proposed an approach based on Q-learning to address (a \textit{centralized} variant of) the SCIM problem consisting of a linear supply chain with four participants.
In particular, they defined the current system state as a vector consisting of the four inventory positions in terms of current stock levels.
However, considering that inventory positions thus defined may take infinite values, applying this strategy appears unfeasible since the Q-table would be in turn infinite.
Consequently, the authors \textit{discretized} the state space into nine intervals.
In this way, the possible state values amount to $9^4$.
Regarding actions, their approach determines the number of products to order via the $d{+}x$ policy;
precisely, if a participant in the previous time step received a request for $d$ product units from the succeeding stage, the \textit{$d{+}x$ policy} requires ordering $d{+}x$ units to the preceding stage in the current time step.
The learning process's objective is hence to determine the value of the unknown variable $x$ according to the given system state.
For limiting the Q-table size, $x$ was \textit{constrained} by the authors to belong to [0, 3] so that the possible number of actions amounts to $4^4$.

Obviously, by defining restricted state and action spaces, the resulting Q-table appears to be more manageable.
However, analyzing various RL studies \cite{Yan2022}, it becomes evident that the Q-tables implemented are typically huge and, thus, \textit{unscalable}.
For example, the Q-table adopted by \cite{Chaharsooghi2008} has a number of cells equal to ($9^4 \cdot 4^4=$) \numprint{1679616}, equivalent to the number of states multiplied by the number of actions.
Consequently, expanding the size of the state or action spaces might not be feasible, as the Q-tables can no longer be handled.

Consequently, tabular RL methods can only be applied to discretized or constrained state and action spaces.
However, discretization leads to a \textit{loss of crucial information}, in addition to being unsuitable for real-world scenarios; 
thus, we need improved RL methods to address the SCIM problem effectively.

In this respect, deep reinforcement learning is a combination of RL with deep learning (DL) which promises to scale to previously intractable decision-making problems, i.e., environments with high dimensional state and action spaces.
DL is rooted into artificial neural networks (ANNs) \cite{LeCun2015},
which are universal approximators capable of providing an optimal approximation of \textit{highly nonlinear functions}.
In practice, function parameters $\theta$ are adjusted during the learning process in order to maximize the expected return (or, alternatively, to minimize the TD error).

The DRL algorithms we implemented belong to the policy-based methods, which can learn a \textit{parameterized} and stochastic policy, $\pi_{\theta} \approx \pi_{*}$ with $\pi: \mathcal{A} \times \mathcal{S} \rightarrow[0,1]$, to select actions directly (as opposed to the Q-learning algorithm, which is part of value-based methods \cite{sutton2018reinforcement}).
Inside them, policy gradient methods offer a considerable theoretical advantage through the \textit{policy gradient theorem}, and the vanilla policy gradient (VPG) algorithm \cite{Williams1992} is a natural result of this theorem;
however, the \textit{high variance} of gradient estimates usually results in policy update instabilities \cite{wu2018variance}.
Also to mitigate this issue, \cite{schulman2015trust} proposed an actor-critic algorithm (which means that a policy and a value function are simultaneously learned) called trust region policy optimization (TRPO), which bounds the difference between the new and the old policy in a \textit{trust region}.
Proximal policy optimization (PPO) \cite{schulman2017proximal} shares the same background as TRPO, but has demonstrated comparable or superior performance while being significantly \textit{simpler} to implement and tune. 
Asynchronous advantage actor-critic (A3C) \cite{mnih2016asynchronous} is also one of the available state-of-the-art actor-critic algorithms.
Its core idea is to have different agents interacting with different representations of the environment, each with its parameters.
Periodically (and asynchronously), they update a global ANN that incorporates \textit{shared parameters}.
For interested readers, an in-depth and more rigorous discussion on the various DRL algorithms can be found in \cite{FranoisLavet2018}.

To the best of the authors' knowledge, only few papers have implemented DRL algorithms to solve the SCIM problem, despite some restrictions.
More in detail, an \textit{extension} of deep Q-network (DQN) \cite{Mnih2015} has been proposed in \cite{Oroojlooyjadid2022} to solve (a \textit{decentralized} variant of) the SCIM problem.
The authors revealed that a DQN agent, which basically involves an ANN instead of a Q-table to return the Q-value for a state-action pair, can learn a near-optimal policy when other supply chain participants follow a base-stock policy;
under a \textit{base-stock policy}, each participant orders in each time step $t$ a quantity to bring its stocks equal to a fixed number $s$, known as the base-stock level, to determine in an optimal way.
Because DQN requires a restricted action space cardinality, the authors performed numerical experiments using a $d{+}x$ policy, with $x$ constrained to one of the following intervals: [-2, +2], [-5, +5], and [-8, +8].

Alternatively, authors in \cite{Peng2019} proposed the VPG algorithm to address a two-echelon supply chain with stochastic and seasonal demand.
Due to storage capacity constraints, the authors designed a \textit{dynamic action space}.
As a result, the number of products to ship is determined also by considering the number of stocks actually present in the warehouses.
To evaluate the VPG performance, three different numerical experiments are presented, and the results show that the VPG agent is able to outperform the ($s$, $Q$)-policy employed as a baseline in all three experiments.
In this context, the \textit{($s$, $Q$)-policy} can be expressed by a rule: at each time step $t$, the current stock level is compared to the reorder point $s$.
If the stock level falls below the reorder point $s$, then the ($s$, $Q$)-policy orders $Q$ units of product;
otherwise, it does not take any action.
Also in this case, the parameters $s$ and $Q$ are to be determined optimally.

Using the same supply chain structure but with ten warehouses and a normal distribution, authors in \cite{Gijsbrechts2022} applied and tuned the A3C algorithm for two different numerical experiments.
The authors restricted the action space by implementing a \textit{state-dependent} base-stock policy, and the results show that A3C can achieve performance comparable to state-of-the-art heuristics and approximate dynamic programming algorithms, despite its initial tuning remaining computationally intensive.

Finally, in the experimental scenario analyzed by \cite{Alves2020}, a general four-echelon supply chain with two nodes per echelon is presented.
The system state consists of product quantity currently available and in transit across the supply chain, plus \textit{future} customer demands.
To deal with the optimization problem, the authors proposed the PPO algorithm, while a deterministic linear programming agent (i.e., considering a deterministic demand) is employed as a baseline.
Results of numerical experiments show that PPO still achieves satisfactory results.

\section{Problem Definition} \label{sec:section4}
\begin{figure}[ht]
    \centering
    \includegraphics[width=0.65\textwidth]{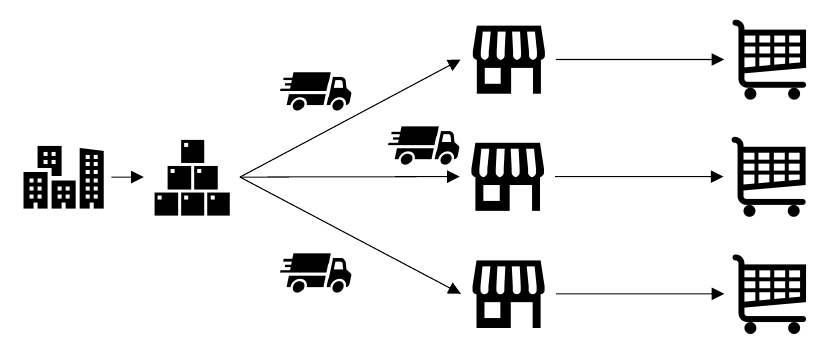}
    \caption{A divergent two-echelon supply chain consisting of a factory and its warehouse (first echelon), plus three distribution warehouses (second echelon).
    Shopping carts represent customers' demands.}
    \label{fig:sc}
\end{figure}

The SCIM environment we propose is primarily motivated by what was presented and discussed in \cite{kemmer2018reinforcement,Peng2019}.
Inspired by these works, we designed a divergent two-echelon supply chain that includes a \textit{factory} that can produce various \textit{product types}, a \textit{factory warehouse}, and a certain number of \textit{distribution warehouses};
an example of this structure is shown in \cref{fig:sc}.

In our formulation, we assume that the factory produces $I$ different product types.
For each product type $i$, the factory decides, at every time step $t$, its respective production level $a_{i, 0, t}$ (we assume $j=0$ for the factory and $1 \leq j \leq J$ for the distribution warehouses), that is, how many units to produce, considering a fixed production cost of $z_{i, 0}$ per unit.
Moreover, the factory warehouse is associated with a maximum capacity of $c_{i, 0}$ units for each product type $i$ (this means that the overall capacity is given by $\sum_{i=0}^{I} c_{i, 0}=c_{0}$).
The cost of storing one unit of product type $i$ at the factory warehouse is $z_{i, 0}^{S}$ per time step, while the corresponding stock level at time $t$ equals $q_{i, 0, t}$.
At every time step $t$, $a_{i, j, t}$ units of product type $i$ are shipped from the factory warehouse to the distribution warehouse $j$, with an associated transportation cost of $z_{i, j}^{T}$ per unit.
For each product type $i$, each distribution warehouse $j$ has a maximum capacity of $c_{i, j}$ ($\sum_{i=0}^{I} c_{i, j}=c_{j}$), a storage cost of $z_{i, j}^{S}$ per unit, and a stock level at time $t$ equal to $q_{i, j, t}$.
The demand for product type $i$ at distribution warehouse $j$ for time step $t$ is equivalent to $d_{i, j, t}$ units, while each unit of product type $i$ is sold to customers at sale price $p_{i}$ (which is identical across all warehouses).

Products are non-perishable and provided in discrete quantities.
Additionally, we assume that each warehouse is legally obligated to fulfill all the submitted orders.
Consequently, if an order for a certain time step exceeds the corresponding stock level, a penalty cost per unsatisfied unit is applied (the penalty cost for product type $i$ is obtained by multiplying the penalty coefficient $z_{i}^{P}$ by the sale price value $p_{i}$).
Unsatisfied orders are maintained over time, and we design them as a negative stock level (which corresponds to \textit{backordering});
this also implies that when the penalty coefficient is particularly high (e.g., $z_{i}^{P} \geq 1$), the agent may not be able to generate a positive profit if it causes backlog orders.
Consequently, it should prefer a policy that leads to accumulating stocks in advance in order to pay storage costs rather than penalty costs.

\subsection{Environment Formulation} 
In this subsection, we formalize the RL problem as an MDP.
More precisely, we introduce and define the \textit{main components} of the SCIM environment that we propose in this paper: the state vector, the action vector, and the reward function.

The \textit{state vector} includes all current stock levels for each warehouse and product type, plus the last $\tau$ demand values, and is defined as follows:
\begin{equation*} \label{eq:state}
\mathrm{s}_{t}=\left(q_{0, 0, t}, \ldots, q_{I, J, t}, d_{t-\tau}, \ldots, d_{t-1}\right),
\end{equation*}
where $d_{t-1}=\left(d_{0, 1, t-1}, \ldots, d_{I, J, t-1}\right)$.
It is worth noticing that the actual demand $d_t$ for the current time step $t$ will not be known until the next time step $t+1$.
This implementation choice ensures that the agent may benefit from learning the demand pattern so as to integrate a sort of \textit{demand forecasting} directly into the policy.
Additionally, we include the last demand values in order to enable the agent to have \textit{limited knowledge} about the demand history and, consequently, to gain a basic comprehension of its fluctuations (similar to what was made originally by \cite{kemmer2018reinforcement}).
In our SCIM implementation, the agent can access the demand values of the last five time steps, even if preliminary results suggest that comparable performances are obtained by accessing the last three or four time steps.

Regarding the \textit{action vector}, we chose to implement a \textit{continuous action space} (i.e., the ANN generates the action value directly) consisting, for each product type, of the number of units to produce at the factory and of the number of units to ship to each distribution warehouse:
\begin{equation} \label{eq:action}
\mathrm{a}_{t}=\left(a_{0, 0, t}, \ldots, a_{I, J, t}\right).
\end{equation}
Usually, a relatively small and \textit{identical upper bound} is typically adopted for all the action values to reduce the computational effort.
However, the drawback is that this might lead to a significant drop in terms of performance.
Indeed, if the upper bound is set too small, the agent may select an inefficient action given that the optimal one is outside the admissible range.
Otherwise, if the upper bound is set too high, the agent may repeatedly choose an incoherent action, i.e., one that falls within the admissible range but exceeds a specified maximum capacity, consequently slowing down the training process.

Our implementation thus provides a continuous action space with an \textit{independent upper bound} for each action value, in order to find a trade-off between efficiency and performance.
In practical terms, the lower bound for each value is simply zero.
In fact, it would be illogical to produce or ship negative quantities of products.
Conversely, the upper bound for each distribution warehouse corresponds to its maximum capacity with respect to each product type (by referring to \cref{eq:action}, $0 \leq a_{i, j, t} \leq c_{i, j}$).
To guarantee that the factory can adequately handle the various demands, its upper bound amounts to the sum of all warehouses' capacities with regard to each product type ($0 \leq a_{i, 0, t} \leq \sum_{j=0}^{J} c_{i, j}$).
We expect to improve both efficiency and performance with this intuition, as the action space is bounded (and hence restricted) but contains only coherent (and possibly optimal) actions.
We specify that available stocks are not explicitly considered when the agent chooses an action. 
However, producing or shipping a number of stocks that it is not possible to store leads to a cost and, therefore, an \textit{implicit penalty} for the agent.
We also assume that there are no lead times both for production and transportation (or, to refer to the literature, we consider \textit{constant lead times equal to 0}). 
This assumption allows us to isolate the primary dynamics of the problem without the additional effects of lead times, thus making the problem easier to address and manage.

\begin{figure}[ht]
    \centering
    \begin{subfigure}[b]{0.49\textwidth}
        \includegraphics[width=1\linewidth]{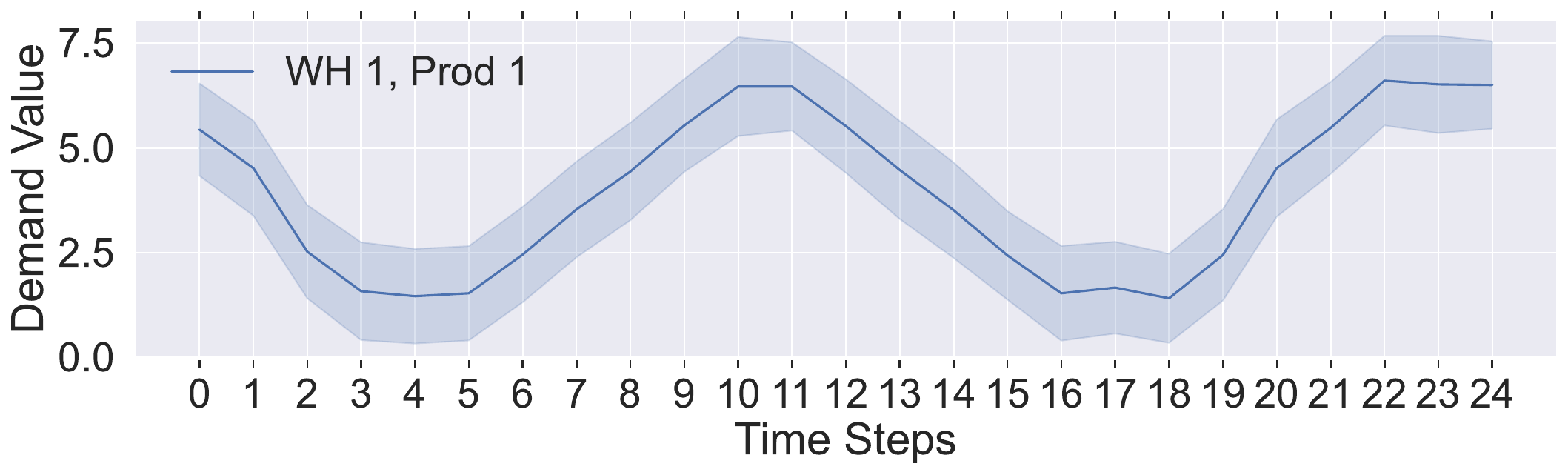}
        \subcaption{}
        \label{fig:1p1wd} 
    \end{subfigure}
    \begin{subfigure}[b]{0.49\textwidth}
        \includegraphics[width=1\linewidth]{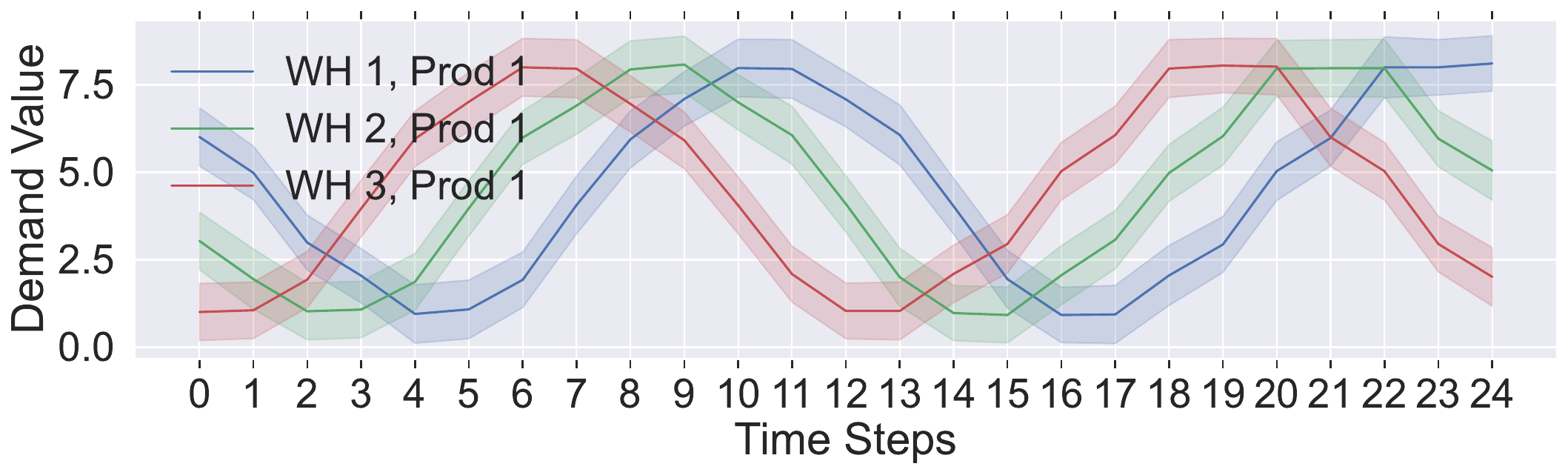}
        \caption{}
        \label{fig:1p3wd} 
    \end{subfigure}
    \begin{subfigure}[b]{0.49\textwidth}
        \includegraphics[width=1\linewidth]{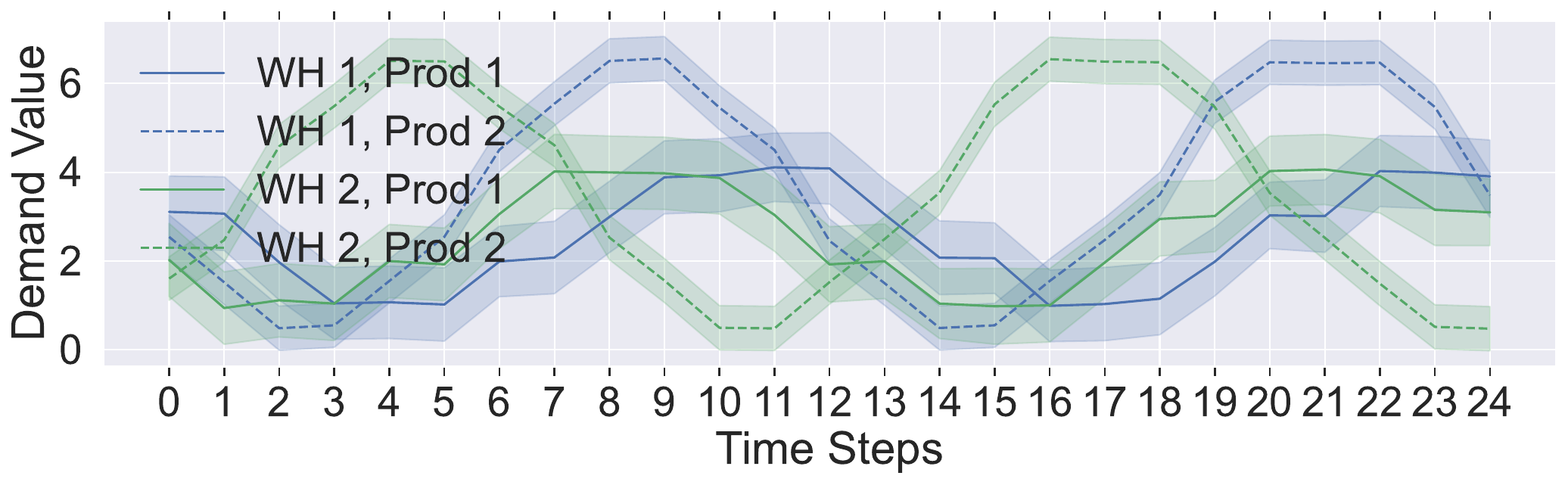}
        \caption{}
        \label{fig:2p2wd} 
    \end{subfigure}
    \caption{Some instances of different demands behavior generated according to \cref{eq:demand} for different topologies and configurations of the SCIM problem: \subref{fig:1p1wd} one product type and one distribution warehouse with $d_{max} = 5$ and $d_{var} = 3$; \subref{fig:1p3wd} one product type and three distribution warehouses with $d_{max} = 7$ and $d_{var} = 2$; and \subref{fig:2p2wd} two product types and two distribution warehouses with $d_{max} = (3, 6)$ and $d_{var} = (2, 1)$ (referring to the values in round brackets, the first denotes the first product type, whereas the second indicates the second product type).}
    \label{fig:demands} 
\end{figure}

To evaluate the performance of the DRL agents, we simulate a seasonal behavior by representing the \textit{demand} as a co-sinusoidal function with a stochastic component, defined according to the following equation: 
\begin{equation} \label{eq:demand}
\begin{aligned}
d_{i, j, t} = &\biggl\lfloor \frac{d_{max_{i}}}{2} \left( 1+\cos \left( \frac{4\pi(2ij+t)}{T} \right)\right) + \mathcal{U}\left(0, d_{var_{i}} \right)\biggr\rfloor,
\end{aligned}
\end{equation}
where $\lfloor\cdot\rfloor$ is the floor function, $d_{max_{i}}$ is the maximum demand value for each product type, $\mathcal{U}$ is a random variable uniformly distributed on the support $(0, d_{var_{i}})$ representing the demand variations (i.e., the \textit{uncertainty}), and $T$ is the final time step of the episode.
At each time step $t$, the demand may vary for each distribution warehouse $j$ and product type $i$ while maintaining the same behavior, as can be seen in \cref{fig:demands}.

The \textit{reward function} for each time step $t$ is then defined as follows:
\begin{equation}\label{eq:reward}
\begin{aligned}
\mathrm{r}_{t}
&= \sum_{j=1}^{J} \sum_{i=0}^{I} p_{i} \cdot d_{i, j, t} -\sum_{i=0}^{I} z_{i, 0} \cdot a_{i, 0, t}  -\sum_{j=1}^{J} \sum_{i=0}^{I} z_{i, j}^{T} \cdot a_{i, j, t} 
\\&-\sum_{j=0}^{J} \sum_{i=0}^{I} z_{i, j}^{S} \cdot \max (q_{i, j, t}, 0) +\sum_{j=0}^{J} \sum_{i=0}^{I} z_{i}^{P} \cdot p_{i} \cdot \min (q_{i, j, t}, 0).
\end{aligned}
\end{equation}
The first term represents revenues, the second one production costs, while the third one transportation costs.
The fourth term is the overall storage costs. 
The function $\max$ is implemented to avoid negative inventories (i.e., backlog orders) from being counted.
The last term denotes the penalty costs, which is introduced with a plus sign because stock levels would already be negative in the eventuality of unsatisfied orders.
The DRL agents' goal is thus to \textit{maximize the supply chain profit} as defined in the reward function.
By design, revenues are always calculated regardless of whether the demand is effectively satisfied;
however, in the event of unsatisfied orders, the penalty costs will impact the actual return for each time step in which backlog orders are counted (in the amount of the penalty coefficient).

Finally, the \textit{state's updating rule} is defined as follows:
\begin{equation*}
\begin{aligned}
s_{t+1}=(&\min [(q_{0, 0, t}+a_{0, 0, t}-\sum_{j=1}^{J} a_{0, j, t}), c_{0, 0}], 
\cdots, 
\\& \min \left[\left(q_{I, J, t}+a_{I, J, t}-d_{I, J, t}\right), c_{I, J}\right], 
d_{t+1-\tau}, 
\cdots, 
d_{t}).
\end{aligned}
\label{eq:next_state}
\end{equation*}
This implies that, at the beginning of the next time step, the factory's stocks are equal to the initial stocks, plus the units produced, minus the stocks shipped.
Similarly, the distribution warehouses' stocks are equal to the initial stocks, plus the units received, minus the current demand.
When surplus stocks are generated, a storage cost is imposed;
otherwise, a penalty cost is considered.
Lastly, the demand values included in the state vector are also updated, discarding the oldest value and concatenating the most recent one.

\section{Numerical Experiments} \label{sec:section5}
Once the environment has been specified, we implemented the agents according to three different state-of-the-art DRL algorithms: A3C, PPO, and VPG, which have been briefly introduced in \cref{sec:section2}.
In this respect, we relied on the implementations made available by Ray \footnote{The Ray library is available on \url{https://www.ray.io}.}, an open-source Python framework that is bundled with RLib, a scalable RL library, and Tune, a scalable hyperparameter tuning library.
An advantage of Ray is that it natively supports OpenAI Gym.
As a result, we exploited the OpenAI Gym APIs to develop the \textit{simulator} representative of the environment and used for the agents' training process.

To assess and compare performances achieved by the adopted DRL algorithms, we also implemented a static reorder policy known in the specialized literature as the ($s$, $Q$)-policy.
In our implementation, we opted to make reordering decisions independently;
this means that the ($s$, $Q$)-policy parameters, $s_{i,j}$ and $Q_{i,j}$, can differ for each warehouse and product type (this policy is still defined \textit{static} because these parameters do not change over time).
To find the best possible parameters that maximize \cref{eq:reward}, we developed a \textit{data-driven approach} based on Bayesian optimization (BO).
In this way, the solution method does not require making any assumptions or simplifications, and hence it is no longer problem-dependent; 
therefore, it can be applied to any SCIM topology and configuration just as it happens for DRL algorithms (they share, in fact, the same identical simulator).

To compare DRL and BO approaches, we also implemented an \textit{oracle}, that is, a baseline that knows the real demand value for each product type and distribution warehouse in advance and can accordingly select the optimal actions to take a priori.

\subsection{Scenarios Considered}
A rich set of numerical experiments have been designed and performed to compare the performances of DRL algorithms and BO under \textit{three different scenarios}.
Each scenario is associated with different demand patterns with respect to each product type and distribution warehouse (i.e., seasonal and stochastic fluctuations).
Furthermore, each scenario has different capacities and costs for evaluating in-depth the adaptability and robustness of DRL algorithms.

Under the \textit{one product type one distribution warehouse} (1P1W) scenario, the supply chain is set to manage just one product type. 
Accordingly, it consists of one factory, a factory warehouse, and one distribution warehouse;
thus the input dimension of the ANN (representing the state vector) is equal to 7, given by the number of warehouses (i.e., 2, including the factory warehouse) times the number of product types (i.e., 1), plus the last demand values for each distribution warehouse and product type (i.e., 5), while the output dimension of the ANN (expressing the action vector) is 2, equivalent to the number of warehouses (including the factory warehouse) multiplied by the number of product types.
Under the 1P1W scenario, which consists of five experiments (as summarized in Table 1 of the supplementary material \footnote{The supplementary material is available on \url{https://github.com/frenkowski/SCIMAI-Gym}.}), sale prices and costs are manipulated so as to increase or decrease revenues and, consequently, the margin of return.
Moreover, in the first experiment, we bound the warehouses' capacities in such a way that they are smaller than the maximum demand value (also considering the stochastic demand variation).
This decision is made to study whether DRL algorithms are able to learn an efficient strategy, i.e., a strategy capable of predicting a \textit{growing demand} and thus saving and shipping stocks in advance.
Analogously, we expect a greater quantity of stocks to be stored and shipped when storage and transportation costs are low, while we expect the opposite when these costs are high.
Finally, we generate multiple penalty coefficients to determine whether a hefty punishment forces DRL algorithms to be more or less effective, with particular attention to the more challenging experiments where low revenues and high costs are considered.

The \textit{one product type three distribution warehouses} (1P3W) scenario concerns a more complex configuration, consisting of a factory, a factory warehouse, and three distribution warehouses. 
Even in this case, the supply chain still manages a single product type, while the input and output dimensions of the ANN are equal to 9 and 4, respectively;
hence, the difficulty of the problem is increased because there is a higher number of both ANN parameters to be optimized and actions to be determined.
The design of the five experiments follows that of the previous 1P1W scenario.
However, a remarkable difference is found in storage capacities and costs (as depicted in Table 2 of the supplementary material \footnotemark[8]).
In fact, we set warehouses' costs to be \textit{directly proportional} to their corresponding capacities, that is, the less storage space we have, the more expensive it is to store a product.
This scenario is also designed to investigate the DRL algorithms strategy when capacities increase, given that the search space of optimal actions grows accordingly.
Furthermore, we are interested in studying how DRL algorithms react when demand, with the associated costs (i.e., production and transportation), becomes greater than actual capacities, considering that the supply chain now consists of three distribution warehouses and, consequently, the SCIM problem becomes more challenging to be tackled.

Finally, in the \textit{two product types two distribution warehouses} (2P2W) scenario, the supply chain consists of two product types, a factory with its warehouse, and two distribution warehouses.
With this design, the number of parameters to optimize is still higher, considering that the ANN input dimension is equal to 26, while the ANN output dimension is 6.
Due to computational time, we performed just three experiments under this scenario (as reported in Table 3 of the supplementary material \footnotemark[8]).
Regarding the demand, we explore demand variations which can be different or equal, according to the specific experiment.
Additionally, we thought of something different concerning storage capacities and, consequently, the search space of optimal actions.
Indeed, in the last experiment, warehouses' capacities for the first product type are designed in descending order, while for the second product type in ascending order;
this implies that, for example, the second distribution warehouse can store the minimum amount of stocks for the first product type and the maximum amount for the second product type.
We expect that this \textit{imbalance}, especially when combined with greater uncertainty, makes the SCIM problem more unexpected and, thus, more difficult to be effectively solved.

\section{Results} \label{sec:section6}
To compare the performances between DRL algorithms, BO, and oracle, we simulated, for each scenario and experiment, 200 different episodes.
Each episode consists of 25 time steps, and we reported the \textit{average cumulative profit} achieved, i.e., the sum of the per-step profit at the last time step $T$.
All experiments were run on a machine equipped with an Intel\textsuperscript{\textregistered} Xeon\textsuperscript{\textregistered} Platinum 8272CL CPU at 2.6 GHz and 16 GB of RAM.
The hyperparameters of DRL algorithms selected for tuning have been chosen following what is presented in the Ray documentation and discussed in the papers \cite{Alves2020,Gijsbrechts2022} (they are reported in Table 4 of the supplementary material \footnotemark[8], along with their corresponding values).
To early stop training instances associated with \textit{bad hyperparameters configurations}, we also implemented, through Ray, the asynchronous successive halving (ASHA) scheduling algorithm \cite{li2018system}.
It is important to note that the simulation results presented and commented in this section have been obtained by selecting, for each algorithm and experiment, the respective \textit{best training instance} \footnote{All the figures regarding the three scenario and related to the convergence and the behavior of DRL algorithms and BO are available on \url{https://github.com/frenkowski/SCIMAI-Gym}.}.

Results of numerical experiments under the 1P1W scenario are summarized in \cref{tab:result1}.
BO and PPO achieve a near-optimal profit in the first experiment where the demand is greater than warehouses' capacities, whereas A3C and VPG perform slightly worse.
All DRL algorithms achieve comparable results in the second and simpler experiment, with higher revenues but lower transportation and penalty costs.
In the third and more complex experiment, which, on the contrary, involves lower revenues and higher transportation costs and penalties, the optimal profit is relatively small, but PPO tends to behave better than other DRL algorithms.
BO, PPO, and A3C obtain satisfactory profits in the fourth and more balanced experiment, with increasing revenues and maximum demand value but reducing uncertainty, while VPG seems to perform poorly.
The main difficulty here is represented by a wider search space (caused by greater storage capacities) and higher storage costs, especially for the factory.
In the fifth and last experiment, the demand uncertainty increases, the penalty costs decrease, and it is more expensive to maintain stocks at the distribution warehouse rather than at the factory, but all DRL algorithms achieve comparable and near-optimal results.

\cref{tab:result2} summarizes the results for the 1P3W scenario, which in design is similar to the 1P1W scenario.
The first experiment is characterized by a high maximum demand value, especially if compared with the capacities of the factory and of the first distribution warehouse;
with this setting, BO performs worse than DRL algorithms.
However, as PPO, it obtains a nearly optimal profit in the second experiment, where a simpler configuration is investigated.
In the third and more challenging experiment, none of the algorithms achieves a profit greater than zero, with PPO achieving the worst one.
Still, PPO outperforms A3C and VPG in the fourth and more balanced experiment, characterized by an increased search space and higher storage costs. 
Finally, BO and PPO achieve the best profits in the fifth experiment, where uncertainty and search space are increased, but fewer penalties are considered.

To conclude, \cref{tab:result3} summarizes performances under the 2P2W scenario.
The first experiment provides a balanced configuration, with maximum demand values and variations that change according to the specific product type, storage costs at the factory greater than those at the two distribution warehouses, and revenues particularly high for the first product type.
Under such a mix, PPO achieves a good profit, as it also does A3C, which overcomes BO.
For the second experiment, sales prices for the second product type are increased and, accordingly, the associated revenues grow as well.
Even storage and transportation costs are decreased, while penalties increase.
With this configuration, PPO still obtains a nearly optimal result, and the same happens for VPG, while BO also behaves well.
In the third experiment, capacities are increased, and we design alternating storage costs;
this means, for example, that maintaining stocks of the first product type at the factory warehouse is the most inexpensive option while maintaining stocks of the second product type is the most expensive.
The results allow us to conclude that PPO, followed by VPG, continues to perform successfully, whereas BO seems to suffer the most.

\begin{table*}[hb!]
\caption{Results related to the three scenarios considered:
\subref{tab:result1} for the 1P1W scenario, it is possible to note how BO and PPO obtain near-optimal profits in general, while A3C and VPG seem more distant in terms of performance;
\subref{tab:result2} in the 1P3W scenario, PPO performs better than BO and other DRL algorithms on average, except in the third and more challenging experiment;
\subref{tab:result3} results concerning the 2P2W scenario suggest that PPO behaves well typically, whereas BO seems slightly inferior compared to the other DRL algorithms.}

\begin{subtable}[b]{\textwidth}
\centering
\caption{} 
\footnotesize
\begin{tabularx}{\textwidth}{@{\extracolsep{\fill}} l D{.}{\;\pm\;}{5.4} D{.}{\;\pm\;}{5.4} D{.}{\;\pm\;}{5.4} D{.}{\;\pm\;}{5.4} | D{.}{\;\pm\;}{5.4} @{\extracolsep{\fill}}}
\toprule
 & \multicolumn{1}{c}{\textbf{A3C}} & \multicolumn{1}{c}{\textbf{PPO}} & \multicolumn{1}{c}{\textbf{VPG}} & \multicolumn{1}{c|}{\textbf{BO}} & \multicolumn{1}{c}{\textbf{Oracle}} \\ \midrule
\textbf{Exp 1} & 870.67  & 1213.68 & 885.66  & \textbf{1226}.\textbf{71} & 1474.45 \\
\textbf{Exp 2} & 1066.94 & 1163.66 & 1100.77 & \textbf{1224}.\textbf{60} & 1289.68 \\
\textbf{Exp 3} & -36.74  & \textbf{195}.\textbf{43}  & 12.61   & 101.50  & 345.18  \\
\textbf{Exp 4} & 1317.60 & 1600.62 & 883.95  & \textbf{1633}.\textbf{39} & 2046.37 \\
\textbf{Exp 5} & 736.45  & 838.58  & 789.51  & \textbf{870}.\textbf{67}  & 966.55  \\ \bottomrule
\end{tabularx}
\label{tab:result1}
\end{subtable}

\begin{subtable}[b]{\textwidth}
\centering
\caption{} 
\footnotesize
\begin{tabularx}{\textwidth}{@{\extracolsep{\fill}} l D{.}{\;\pm\;}{5.4} D{.}{\;\pm\;}{5.4} D{.}{\;\pm\;}{5.4} D{.}{\;\pm\;}{5.4} | D{.}{\;\pm\;}{5.4} @{\extracolsep{\fill}}}
\toprule
 & \multicolumn{1}{c}{\textbf{A3C}} & \multicolumn{1}{c}{\textbf{PPO}} & \multicolumn{1}{c}{\textbf{VPG}} & \multicolumn{1}{c|}{\textbf{BO}} & \multicolumn{1}{c}{\textbf{Oracle}} \\ \midrule
\textbf{Exp 1} & 1606.139  & \textbf{2319}.\textbf{122}  & 803.154   & 486.330   & 3211.60 \\
\textbf{Exp 2} & 2196.104  & \textbf{3461}.\textbf{120}  & 2568.112  & 3193.101  & 3848.95 \\
\textbf{Exp 3} & -2142.128 & -4337.216 & -2638.121 & \textbf{-1682}.\textbf{196} & 772.21  \\
\textbf{Exp 4} & -561.237  & \textbf{2945}.\textbf{135}  & 656.140   & 1256.170  & 4389.64 \\
\textbf{Exp 5} & 1799.306  & \textbf{2353}.\textbf{131}  & 1341.79   & 2203.152  & 2783.91 \\ \bottomrule
\end{tabularx}
\label{tab:result2}
\end{subtable}

\begin{subtable}[b]{\textwidth}
\centering
\caption{}
\footnotesize
\begin{tabularx}{\textwidth}{@{\extracolsep{\fill}} l D{.}{\;\pm\;}{5.4} D{.}{\;\pm\;}{5.4} D{.}{\;\pm\;}{5.4} D{.}{\;\pm\;}{5.4} | D{.}{\;\pm\;}{5.4} @{\extracolsep{\fill}}}
\toprule
 & \multicolumn{1}{c}{\textbf{A3C}} & \multicolumn{1}{c}{\textbf{PPO}} & \multicolumn{1}{c}{\textbf{VPG}} & \multicolumn{1}{c|}{\textbf{BO}} & \multicolumn{1}{c}{\textbf{Oracle}} \\ \midrule
\textbf{Exp 1} & 2227.178 & \textbf{2783}.\textbf{139} & 1585.184 & 2086.173 & 3787.102 \\
\textbf{Exp 2} & 1751.83  & \textbf{2867}.\textbf{90}  & 2329.98  & 2246.114 & 3488.63  \\
\textbf{Exp 3} & 1414.128 & \textbf{2630}.\textbf{138} & 2434.156 & 552.268  & 3549.103 \\ \bottomrule
\end{tabularx}
\label{tab:result3}
\end{subtable}
\end{table*}

\section{Discussions and Conclusions} \label{sec:section7}
Results of numerical experiments demonstrated that the SCIM environment we propose is \textit{effective} in representing states, actions, and rewards;
indeed, the DRL algorithms we implemented have achieved \textit{nearly optimal solutions} in all three investigated scenarios.
In detail, PPO is the one that better adapts to different topologies and configurations of the SCIM environment achieving higher profits than other algorithms on average, although it fails to reach a positive profit in the most challenging experiment of the 1P3W scenario.
VPG frequently appears to converge to a local maximum that seems slightly distant from PPO, especially when the number of warehouses increases, but it still obtains acceptable results.

It is worthwhile to mention that the BO approach also shows remarkable results, especially when the search space of optimal actions is limited, as in the 1P1W scenario.
When compared to DRL algorithms, the BO approach seems to suffer more when there are two product types or when the demand exceeds the capacities.
This is mainly due to the static and non-dynamic nature of the ($s$, $Q$)-policy, which does not allow developing an effective strategy, for example, for saving stocks in advance, but, conversely, culminates in a \textit{myopic behavior}.
Nevertheless, the absence of hyperparameters to be tuned offers a considerable advantage.

\subsection{Future Research} 
This paper can be extended and improved in many directions as: 
\begin{itemize}
    \item Develop a \textit{more comprehensive SCIM environment}, for example, by considering additional configurations mentioned in \cite{deKok2018} (e.g., different demand distributions or different customers' reactions to backordering).
    \item Take into account the \textit{non-linearity of transportation costs} (e.g., introducing a fixed cost independent of the number of stocks shipped effectively), as well as \textit{non-zero leading times}.
    \item Use \textit{real-world data} to validate DRL algorithms and check whether they improve the performances of currently used SCIM systems in practice.
\end{itemize}

Lastly, even the BO approach could be \textit{extended} to other standard static reorder policies, such as the base-stock policy, which has exactly half of the ($s$, $Q$)-policy parameters and can therefore enable faster convergence times.

%\subsubsection{Acknowledgements} Please place your acknowledgments at
%the end of the paper, preceded by an unnumbered run-in heading (i.e.
%3rd-level heading).

\bibliographystyle{splncs04}
\bibliography{mybibliography}

\end{document}

\title{Comparing Deep Reinforcement Learning Algorithms in Two-Echelon Supply Chains}
\titlerunning{Comparing DRL Algorithms in Two-Echelon Supply Chains}
\author{Francesco Stranieri\thanks{Corresponding author. E-mail: \email{francesco.stranieri@polito.it}.}\inst{1,2}\orcidID{0000-0002-5366-8499} \and
\\Fabio Stella\inst{1}\orcidID{0000-0002-1394-0507}}
\authorrunning{F. Stranieri and F. Stella}
\institute{University of Milan-Bicocca, Milan MI 20125, Italy \and
Polytechnic of Turin, Turin TO 10129, Italy}
\maketitle              % typeset the header of the contribution
\begin{abstract}
In this study, we analyze and compare the performance of state-of-the-art deep reinforcement learning algorithms for solving the supply chain inventory management problem. 
This complex sequential decision-making problem consists of determining the optimal quantity of products to be produced and shipped across different warehouses over a given time horizon.
In particular, we present a mathematical formulation of a two-echelon supply chain environment with stochastic and seasonal demand, which allows managing an arbitrary number of warehouses and product types.
Through a rich set of numerical experiments, we compare the performance of different deep reinforcement learning algorithms under various supply chain structures, topologies, demands, capacities, and costs.
The results of the experimental plan indicate that deep reinforcement learning algorithms outperform traditional inventory management strategies, such as the static (s, Q)-policy.
Furthermore, this study provides detailed insight into the design and development of an open-source software library that provides a customizable environment for solving the supply chain inventory management problem using a wide range of data-driven approaches.

\keywords{artificial intelligence \and deep learning \and reinforcement learning \and smart manufacturing \and inventory management.}
\end{abstract}
\section*{Supplementary Material} \label{sec:sup_mat}

\clearpage

\appendix
%\crefalias{section}{appendix}
\section{Experimental Plan} \label{sec:exp_plan}

\begin{table*}[ht]
\caption{Experiments concerning the three scenarios considered:
\subref{tab:scenario1} for the 1P1W scenario, when two values are present, the first one refers to the factory, while the second one refers to the first (and only) distribution warehouse.;
\subref{tab:scenario2} for the 1P3W scenario, when there are four values, the first relates to the factory, while the remaining to the first, the second, and the third distribution warehouse, respectively;
\subref{tab:scenario3} referring to the round brackets of the 2P2W scenario, the first value denotes the first product type, whereas the second value indicates the second product type.}

\begin{subtable}[ht]{\textwidth}
\centering
\caption{}
\tiny
\begin{tabularx}{\textwidth}{@{\extracolsep{\fill}} lccccc X@{}}
\toprule
                              & \textbf{Exp 1} & \textbf{Exp 2} & \textbf{Exp 3} & \textbf{Exp 4} & \textbf{Exp 5} \\ \midrule
\textbf{Max Demand Value}     & 10             & 5              & 5              & 10             & 5              \\
\textbf{Max Demand Variation}       & 2              & 2              & 2              & 1              & 3              \\
\textbf{Sale Price}           & 15             & 20             & 15             & 20             & 15             \\
\textbf{Production Cost}      & 5              & 5              & 10             & 5              & 5              \\
\textbf{Storage Capacities}   & 5, 10          & 5, 10          & 5, 10          & 10, 15         & 5, 10          \\
\textbf{Storage Costs}        & 2, 1           & 2, 1           & 2, 1           & 4, 2           & 1, 2           \\
\textbf{Transportation Cost}  & 0.25           & 0.05           & 1              & 0.25           & 0.25           \\
\textbf{Penalty Coefficient}         & 1.5            & 0.1            & 2              & 1.5            & 0.1            \\ \bottomrule
\end{tabularx}%
\label{tab:scenario1}
\end{subtable}

\begin{subtable}[t]{\textwidth}
\centering
\caption{}
\tiny
\begin{tabularx}{\textwidth}{@{\extracolsep{\fill}} lccccc X@{}}
\toprule
\textbf{}                     & \textbf{Exp 1} & \textbf{Exp 2}   & \textbf{Exp 3} & \textbf{Exp 4} & \textbf{Exp 5} \\ \midrule
\textbf{Max Demand Value}     & 7   & 5   & 5  & 7   & 5   \\
\textbf{Max Demand Variation} & 2   & 2   & 2  & 1   & 3   \\
\textbf{Sale Price}      & 15  & 20  & 15 & 20  & 15  \\
\textbf{Production Cost}           & 5   & 5   & 10 & 5   & 5   \\
\textbf{Storage Capacities}   & 3, 6, 9, 12    & 3, 6, 9, 12      & 3, 6, 9, 12    & 4, 8, 12, 16   & 4, 8, 12, 16   \\
\textbf{Storage Costs}        & 4, 3, 2, 1     & 4, 3, 2, 1       & 4, 3, 2, 1     & 8, 6, 4, 2     & 4, 3, 2, 1     \\
\textbf{Transportation Costs} & 0.3, 0.6, 0.9  & 0.03, 0.06, 0.09 & 3, 2, 1        & 0.3, 0.6, 0.9  & 0.3, 0.6, 0.9  \\
\textbf{Penalty Coefficient}         & 1.5 & 0.1 & 2  & 1.5 & 0.1 \\ \bottomrule
\end{tabularx}%
\label{tab:scenario2}
\end{subtable}

\begin{subtable}[t]{\textwidth}
\centering
\caption{}
\tiny
\begin{tabularx}{\textwidth}{@{\extracolsep{\fill}} lccc X@{}}
\toprule
                               & \textbf{Exp 1} & \textbf{Exp 2} & \textbf{Exp 3} \\ \midrule
\textbf{Max Demand Values}     & 3, 6           & 3, 6           & 4, 2           \\
\textbf{Max Demand Variations} & 2, 1           & 2, 1           & 2, 2           \\
\textbf{Sale Prices}      & 20, 10         & 10, 15         & 20, 10         \\
\textbf{Production Costs}           & 2, 1           & 2, 1           & 2, 1           \\
\textbf{Storage Capacities}    & (3, 4), (6, 8), (9, 12) & (3, 4), (6, 8), (9, 12)            & (9, 4), (6, 8), (3, 12) \\
\textbf{Storage Costs}        & (6, 3), (4, 2), (2, 1)  & (0.5, 0.3), (1.0, 0.6), (1.5, 0.9) & (1, 3), (2, 2), (3, 1)  \\
\textbf{Transportation Costs} & (0.1, 0.3), (0.2, 0.6)  & (0.01, 0.025), (0.02, 0.050)       & (0.1, 0.3), (0.2, 0.6)  \\
\textbf{Penalty Coefficient}          & 0.5            & 1.5            & 0.5            \\ \bottomrule
\end{tabularx}%
\label{tab:scenario3}
\end{subtable}
\end{table*}

\clearpage

\section{Hyperparameters Tuning} \label{sec:hyper_tuning}

\begin{table*}[ht]
\centering
\caption{The hyperparameters of DRL algorithms selected for tuning.
Through a grid search, we instantiated \numprint{880} DRL algorithm instances for the 1P1W scenario, \numprint{880} for the 1P3W, and \numprint{528} for the 2P2W, for a total of \numprint{2288} instances.
Each instance is trained for a given number of episodes: \numprint{15000} episodes for the 1P1W scenario and \numprint{50000} for the 1P3W and 2P2W scenarios.}
\tiny
\begin{tabularx}{\textwidth}{@{\extracolsep{\fill}} lccc X@{}}
\toprule
                                 & \textbf{A3C} & \textbf{PPO} & \textbf{VPG} \\ \midrule
\textbf{Hidden Layers} & \{(64, 64), (128, 128)\} & \{(64, 64), (128, 128)\} & \{(64, 64), (128, 128)\} \\
\textbf{Learning Rate}           & \{1e-4, 1e-3\}   & \{5e-4, 5e-3\}   & \{4e-4, 4e-3\}   \\
\textbf{Rollout Fragment Length} & \{10, 100\}      & \{20, 200\}      & \{10, 100\}      \\
\textbf{Train Batch Size}        & \{200, 2000\}    & \{400, 4000\}    & \{200, 2000\}    \\
\textbf{Grad Clip}               & \{20, 40\}       & \{0, 20\}        & -            \\
\textbf{SGD Mini-Batch Size}     & -            & \{128, 256\}     & -            \\
\textbf{SGD Iterations}          & -            & \{15, 30\}       & -            \\ \bottomrule
\end{tabularx}%
\label{tab:hyper}
\end{table*}